%% file: main.tex
\documentclass[sigconf]{acmart}
\usepackage{multirow}

\usepackage{colortbl}

\usepackage[normalem]{ulem}
\useunder{\uline}{\ul}{}
\usepackage{caption}
\usepackage{subcaption}
\newcommand{\algo}{DELSumm }
\newcommand{\algoname}{\textbf{D}omain-adaptive \textbf{E}xtractive \textbf{L}egal \textbf{Summ}arizer }

\renewcommand\footnotetextcopyrightpermission[1]{} 

\makeatletter
\def\@copyrightspace{\relax}
\makeatother

\begin{document}

\title{Incorporating Domain Knowledge for Extractive Summarization of Legal Case Documents}

\author{Paheli Bhattacharya}
\affiliation{\institution{Department of CSE, IIT Kharagpur}\country{India}}

\author{Soham Poddar}
\affiliation{\institution{Department of CSE, IIT Kharagpur}\country{India}}

\author{Koustav Rudra}
\affiliation{\institution{L3S Research Center, Leibniz
University, Hannover}\country{Germany}}

\author{Kripabandhu Ghosh}
\affiliation{\institution{Department of CDS, IISER, Kolkata}\country{India}}

\author{Saptarshi Ghosh}
\affiliation{\institution{Department of CSE, IIT Kharagpur}\country{India}}

\begin{abstract}
\input{sections/abstract}
\end{abstract}

\keywords{Legal document summarization; Integer Linear Programming }

\maketitle

\section{Introduction} \label{sec:intro}
\input{sections/introduction}

\section{Related Work} \label{sec:related}
\input{sections/related-work}

\section{Dataset} 
\input{sections/dataset}

\section{Proposed Approach: \algo} \label{sec:algo}
\input{sections/proposed-approach}

\section{Applying \algo on Indian case documents} \label{sec:india-expts}
\input{sections/guidelines-india}

\section{Baselines and Experimental Setup}  
\label{sec:setup}
\input{sections/baselines}

\section{Results and Analyses }
\input{sections/results-india}

\vspace{-2mm}
\section{Using \algo with Algorithmic Rhetorical Labels}
\input{sections/noisy-labels}

\label{sec:noisy-label}

\section{Conclusion} \label{sec:conclu}
\input{sections/conclusion}

\vspace{3mm}
\noindent {\bf Acknowledgements:} 
The authors thank the Law experts from the Rajiv Gandhi School of Intellectual Property Law, India who helped in developing the gold standard data and provided the guidelines for summarization. 
The research is partially supported by 
the TCG Centres for Research and Education in Science and Technology (CREST)
through the project titled `Smart Legal Consultant: AI-based Legal Analytics'. This work is also supported in part by the European Union’s Horizon 2020 research and innovation programme under grant agreement No 832921.
P. Bhattacharya is supported by a Fellowship from Tata Consultancy Services.

\bibliographystyle{ACM-Reference-Format}
\bibliography{references}

\end{document}

%% file: sections/abstract.tex
Automatic summarization of legal case documents is an important and practical challenge.
Apart from many domain-independent text summarization algorithms that can be used for this purpose, several algorithms have been developed specifically for summarizing legal case documents. 
However, most of the existing algorithms do not systematically incorporate domain knowledge that specifies what information should ideally be present in a legal case document summary. 
To address this gap, we propose an unsupervised summarization algorithm DELSumm which is designed to systematically incorporate guidelines from legal experts into an optimization setup. 
We conduct detailed experiments over case documents from the Indian Supreme Court.
The experiments show that our proposed unsupervised method outperforms several strong baselines in terms of ROUGE scores, including both general summarization algorithms and legal-specific ones. In fact, though our proposed algorithm is {\it unsupervised}, it outperforms several {\it supervised} summarization models that are trained over thousands of document-summary pairs.

%% file: sections/introduction.tex
In a Common Law system (followed in most countries including USA, UK, India, Australia), law practitioners have to go through hundreds of legal case documents to understand how the Court behaved in different legal situations. 
Case documents usually span several tens to hundreds of pages and contain complex text, which makes comprehending them difficult even for legal experts. 
Hence, summaries of case documents often prove beneficial. Existing legal IR systems employ legal attorneys or para-legals to write case summaries, which is an expensive process. Hence, automatic summarization of case documents is a practical and well-known challenge~\cite{polsley2016casesummarizer,zhong2019automatic,farzindar2004letsum,saravanan2006improving,liu2019extracting}. 

Document summarization is broadly of two types -- (i)~{\it extractive}, where important sentences are extracted from the source document and included in the summary ~\cite{Luhn:1958, Erkan:2004, liu2019extracting}, and (ii)~{\it abstractive}, where the model attempts to generate a summary using suitable vocabulary ~\cite{bertsum,nallapati2017summarunner}.
This paper focuses on {\it extractive summarization} since it is more common for legal case documents.

Law practitioners in various countries have a set of guidelines for how to summarize court cases, e.g., see~\cite{how-to-brief-case-cuny,intro-case-briefing-northwestern,case-notes-open-univ-UK} for guidelines on summarizing case documents of the USA and the UK. 
All these guidelines basically advise to consider different {\it rhetorical/thematic segments that are usually present in case documents} (e.g., facts of the case, legal issues being discussed, final judgment, etc.) and then to {\it include certain important parts of each segment into the summary}. 
We got similar advice from law experts in India -- two senior law students and a faculty from the Rajiv Gandhi School of Intellectual Property Law, India -- regarding Indian court case documents (see details in Section~\ref{sec:guidelines-india}).
Apart from focusing on summarizing the case document as a whole, it is also necessary that each of these segments is summarized well and has representations in the final summary. 
Another advantage of including representations of each segment in the summary is that, often law practitioners wish to read only the summary of particular segments (e.g., only the facts of a case, or the legal issues involved in a case).

There are several summarization algorithms specifically meant for legal court case documents~\cite{polsley2016casesummarizer,zhong2019automatic,farzindar2004letsum,saravanan2006improving,liu2019extracting} (details in Section~\ref{sec:related}).
However, most of these algorithms do {\it not} take into account the guidelines from law practitioners. 
Though a few legal domain-specific algorithms recognize the presence of various rhetorical segments in case documents~\cite{farzindar2004letsum,saravanan2006improving}, they also do {\it not} consider expert guidelines in deciding what to include in the summary from which segment.
As a result, these algorithms cannot represent some of the important segments well in the summary. For instance, a recent work~\cite{bhattacharya2019comparative} showed that most existing algorithms do {\it not} represent the {\it final judgement} well in the summary, though this segment is considered very important by law practitioners. 

The above limitations of existing methods motivate us to develop a summarization algorithm that will consider various rhetorical segments in a case document, and then decide which parts from each segment to include in the summary, based on guidelines from law practitioners. Supervised methods which are expected to learn these guidelines automatically would require a large number of expert written summaries for lengthy case judgements, which may not be available for all jurisdictions. Hence we go for an {\it unsupervised} summarization algorithm in this work.

We propose \textbf{\algo} (\algoname), an unsupervised extractive summarization algorithm for legal case documents. 
We formulate the task of summarizing legal case documents as maximizing an Integer Linear Programming (ILP) objective function that aims to maximize the inclusion of the most informative sentences in the summary, and has balanced representations from all the thematic segments, while reducing redundancy.
We demonstrate how \algo can be tuned to operationalize summarization guidelines obtained from law experts over Indian Supreme Court case documents.
Comparison with several baseline methods -- including five legal domain-specific methods and two state-of-the-art deep learning models trained over thousands of document-summary pairs -- suggest that our proposed approach outperforms most existing methods, especially in summarizing the different rhetorical segments present in a case document. 
To summarize, the contributions of this paper are:

\noindent (1) We propose \algo  that incorporates domain knowledge provided by legal experts for summarizing lengthy case documents.\footnote{Implementation publicly available at \url{https://github.com/Law-AI/DELSumm}}

\noindent (2) We perform extensive experiments on documents from  the Supreme Court of India and compare our proposed method with eleven baseline approaches.
We find that \algo outperforms a large number of legal-specific as well as general summarisation methods, including supervised neural models trained over thousands of document-summary pairs.
Especially, \algo achieves {\it much better summarization of the individual rhetorical segments} in a case document, compared to most prior methods.

\noindent (3) We also show that our proposed approach is robust to inaccurate rhetorical labels generated algorithmically. There is a negligible drop in performance when \algo uses rhetorical sentence labels generated by a rhetorical segmentation method (which is more a practical setup), instead of using expert-annotated labels.

To the best of our knowledge, this is the first systematic attempt to computationally model and incorporate legal domain knowledge for summarization of legal case documents. 
Through comparison with 11 methods, including 2 state-of-the-art deep learning methods, we show that an unsupervised algorithm developed by intelligently including domain expertise, can surpass the performance of supervised learning models even when the latter are trained over large training data (which is anyway expensive to obtain in an expert-driven domain such as Law).

%% file: sections/related-work.tex
Extractive text summarization aims to detect \textit{important} sentences from the full document and include them in the summary.
Existing methods for extractive summarization that can be applied for legal document summarization, can be broadly classified into four classes: (i)~Unsupervised domain-independent, (ii)~Unsupervised domain-specific, (iii)~Supervised domain-independent, and (iv)~Supervised domain-specific. 
We briefly describe some popular methods from each of the classes in this section.

\subsection{Unsupervised domain-independent methods} \label{sub:related-udi}
Popular, unsupervised extractive summarization algorithms identify important sentences either by using Frequency-based methods (e.g. Luhn~\cite{Luhn:1958}) or Graph-based methods (e.g. LexRank~\cite{Erkan:2004}). The summary is the top ranked sentences. 
There are algorithms from the family of matrix-factorization, such as LSA~\cite{Gong:2001}. 

Summarization has also been treated as an Integer Linear Programming (ILP) optimization problem. Such methods have been applied for summarizing news documents~\cite{banerjee2015multi} and social media posts~\cite{10.1145/3209978.3210030}. 
Our proposed approach in this work is also based on ILP-based optimization.

\subsection{Unsupervised domain-specific methods} \label{sub:related-uds}
There are several unsupervised methods for extractive summarization specifically designed for legal case documents. 
One of the earliest methods for unsupervised legal document summarization that takes into account the rhetorical structure of a case document was {\bf LetSum}~\cite{farzindar2004letsum}. They consider certain cue-phrases to assign rhetorical/semantic roles to a sentence in the source document. They specifically consider the roles -- Introduction, Context, Juridical Analysis and Conclusion. Sentences are ranked based on their TF-IDF values for estimating their importance. The final summary is generated by taking 10\% from the Introduction, 25\% from Context, 60\% from Juridical Analysis and 5\% from the Conclusion segments.

While LetSum considers TF-IDF to rank sentences, Saravanan et.al.~\cite{saravanan2006improving} use a K-Mixture Model to rank sentences for deciding which sentences to include in the summary. We refer to this work as {\bf KMM} in the rest of the paper. Note that this work also identifies rhetorical roles of sentences (using a graphical model). However, this is a {\it post-summarization step} that is mainly used for displaying the summary in a structured way; the rhetorical roles are {\it not} used for generating the summary.

Another method {\bf CaseSummarizer}~\cite{polsley2016casesummarizer} finds the importance of a sentence based on several factors including its TF-IDF value, the number of dates present in the sentence, number of named entities and whether the sentence is at the start of a section. Sentences are then ranked in order to generate the summary comprising of the top-ranked sentences. 

Zhong et.al.~\cite{zhong2019automatic} creates a template based summary for Board of Veteran Appeals (BVA) decisions from the U.S. Department of Veteran Affairs. The summary contains (i) one sentence from the procedural history (ii) one sentence from issue (iii) one sentence from the service history of the veteran (iv) variable number of Reasoning \& Evidential Support sentences selected using Maximum Margin Relevance (v) one sentence from the conclusion. We refer to this method as \textbf{MMR} in this paper.

\vspace{2mm}
\noindent \textbf{Limitations of the methods:} CaseSummarizer does not assume the presence of any rhetorical role. The other methods -- LetSum, KMM, and MMR -- consider their presence but do {\it not} include them while estimating the importance of sentences for generating the summary. 
Specifically, LetSum and KMM use generic term-distributional models (TF-IDF and K-mixture models respectively) and MMR uses Maximum Margin Relevance. As evident, rhetorical roles do not play any role in finally selecting the sentences for inclusion in the summary. 

We believe that it is more plausible to measure the importance of sentences in each rhetorical segment separately using domain knowledge. In fact, for each segment, the parameters measuring sentence importance can vary. As an example, consider the segment `Statute'. One can say that sentences from this segment that actually contain a reference to a Statute/Act name are important. In contrast, this reasoning will not hold for the segment `Facts' and one may derive a different measure based on the presence of certain Part-of-Speech tags e.g., Nouns (names of people, location, etc.). 
Hence, in the present work, we especially include such domain knowledge along with segment representations in a more systematic way for generating summaries. 

\subsection{Supervised domain-independent methods} \label{sub:related-sdi}

Supervised neural (Deep Learning-based) methods for extractive text summarization treat the task as a binary classification problem, where sentence representations are learnt using a hierarchical encoder. 
A survey of the existing approaches can be found in~\cite{DBLP:journals/corr/abs-1804-04589}. 

Two popular extractive methods are NeuralSum~\cite{cheng2016neural} and SummaRunner~\cite{nallapati2017summarunner}. 
These methods use RNN (Recurrent Neural Network) encoders to learn the sentence representations from scratch. Sentence selection into summary is based on -- content, salience, novelty and absolute and relative position importance. These parameters are also learned in the end-to-end models. 

Recently, pretrained encoders (especially transformer-based models such as BERT~\cite{devlin2018bert}) have gained much popularity. These encoders have already been pretrained using a large amount of open domain data. Given a sentence, they can directly output its sentence representation. These models can be used in a supervised summarization by fine-tuning the last few layers on domain-specific data. 
BERTSUM~\cite{bertsum} is a BERT-based extractive summarization method. Unlike SummaRuNNer, here sentence selection into summary is based on trigram overlap with the currently generated summary. 

\vspace{2mm}
\noindent \textbf{Limitations of the neural methods}: These Deep Learning architectures have been evaluated mostly in the news summarization domain, 
and news documents are much shorter and contain simpler language than legal documents. 
Recent works~\cite{ExtendedSumm, xiao2019extractive} show that these methods do not perform well in summarizing scientific articles from PubMed and arxiv, which are longer sequences. However, it has not been explored how well these neural models would work in summarizing legal case documents; we explore this question for two popular neural summarization models in this paper. 

\subsection{Supervised domain-specific methods} \label{sub:related-sds}

Liu et.al.~\cite{liu2019extracting} have recently developed a supervised learning method for extractive summarization of legal documents. Sentences are represented using handcrafted features like number of words in a sentence, position of the sentence in a document etc. Machine Learning (ML) classifiers (e.g., Decision Tree, MLP, LSTM) are then applied for obtaining the final summary. We refer to this method as {\bf Gist}~\cite{liu2019extracting}. 

\vspace{2mm}
\noindent \textbf{Limitations of the method:} Although this method was developed and applied for Chinese legal case documents,  domain-specific attributes such as rhetorical labels  were {\it not} considered.

\subsection{Rhetorical roles in a legal case document}  
\label{sub:rhet-roles}
A legal case document can be structured into thematic  segments, where each sentence can be labelled with a rhetorical role.  
Note that case documents often do {\it not} implicitly specify these rhetorical roles; there exist algorithms that assign rhetorical roles to the sentences~\cite{bhattacharya2019identification, jurix_japanese, saravanan-etal-2008-automatic}.
Different prior works have considered different sets of rhetorical roles~\cite{farzindar2004letsum, saravanan2006improving,zhong2019automatic,bhattacharya2019identification}. 
It was shown in our prior work~\cite{bhattacharya2019comparative} that a mapping between the different sets of rhetorical roles is possible.  

In this work, we consider a set of {\it eight} rhetorical roles suggested in our prior work~\cite{bhattacharya2019identification}. Briefly, the rhetorical roles are as follows --
\textbf{(i)~Facts}: the events that led to filing the case, 
\textbf{(ii)~Issue}: legal questions/points being discussed, 
\textbf{(iii)~Ruling by Lower Court}: case documents from higher courts (e.g., Supreme Court) can contain decisions delivered at the lower courts (e.g. Tribunal), 
\textbf{(iv)~Precedent}: citations to relevant prior cases,
\textbf{(v)~Statute}: citations to statutory laws that are applicable to the case (e.g., Dowry Prohibition Act, Indian Penal Code),
\textbf{(vi)~Arguments} delivered by the contending parties, 
\textbf{(vii)~Ratio}: rationale based on which the judgment is given, and 
\textbf{(viii)~Final judgement} of the present court.

In this work, we attempt to generate a summary that includes representations of all rhetorical segments in the source document (full text). We assume that the rhetorical labels of every sentence in the source document is already labeled either by legal experts or by applying the method proposed in~\cite{bhattacharya2019identification}.

%% file: sections/dataset.tex
\label{sec:dataset-india}

For evaluation of summarization algorithms, we need a set of source documents (full text) and their gold standard summaries. Additionally, since we propose to use the rhetorical labels of sentences in a source document for generating the summary, we need rhetorical label annotations of the source documents.
Additionally, since the supervised methods Gist, SumaRuNNer and BERTSUM require training over large number of document-summary pairs, we also need such a training set. 
In this section, we describe the training set and the evaluation set (that is actually used for performance evaluation of summarization algorithms).

\vspace{2mm}
\noindent \textbf{Evaluation set}: Our evaluation dataset consists of a set of $50$ Indian Supreme Court case documents, where each sentence is tagged with a rhetorical/semantic label by law experts (out of the rhetorical roles described in Section~\ref{sub:rhet-roles}). This dataset is made available by our prior work~\cite{bhattacharya2019identification}.
We asked two senior law students (from the Rajiv Gandhi School of Intellectual Property Law, one of the most reputed law schools in India) to write summaries for each of these $50$ document. 
They preferred to write extractive summaries of approximately one-third of the length of the documents. We asked the experts to summarize each rhetorical segment separately (so that we can evaluate how well various models summarize the individual sections); only, they preferred to summarize the rhetorical segments `Ratio' and `Precedent' together. 

All the summarization methods have been used to generate summaries for these $50$ documents. These summaries were then uniformly evaluated against the two gold standard summaries written by the law students, using the standard ROUGE scores (details given in later sections). We report the average ROUGE scores over the two sets of gold standard summaries.

\vspace{2mm}
\noindent \textbf{Training set} : 
For training supervised methods (Gist, SumaRuNNer and BERTSUM), we need an additional training dataset consisting of document-summary pairs. 
To this end, we crawled $7,100$ Indian Supreme Court case documents and their {\it headnotes} (short abstractive summaries) from \url{http://www.liiofindia.org/in/cases/cen/INSC/} which is an archive of Indian court cases. These document-summary pairs were used to train the supervised summarization models (details in later sections).
We ensured that there was {\it no overlap} between this training set and the evaluation set of $50$ documents.

Note that the headnotes described above are {\it not} considered to be summaries of sufficiently good quality by our law experts, and this is why the experts preferred to write their own summaries as gold standard for the documents in the evaluation set. 
It can be argued that it is unfair to train the supervised summarization models using summaries that are known to be of poorer quality than the target summaries. 
However, the supervised summarization models require thousands of document-summary pairs for training, and it is practically impossible to obtain so many summaries of the target quality. 
Hence our only option is to use the available headnotes for training the supervised summarization models.
This situation can be thought of as a trade-off between quantity and quality of training data -- if a method requires large amounts of training data, then that data may be of poorer quality than the target quality.

%% file: sections/proposed-approach.tex
In this section, we describe our proposed algorithm \algo. 
Our algorithm uses an optimization framework to incorporate legal domain knowledge (similar to what is stated in~\cite{how-to-brief-case-cuny,intro-case-briefing-northwestern,case-notes-open-univ-UK}) into an objective function with constraints. The objective function is then maximized using Integer Linear Programming (ILP).
The symbols used to explain the algorithm are stated in Table~\ref{tab:symboltable}.

\begin{table}[tb]

\renewcommand{\arraystretch}{1.1}
\small
\begin{tabular}{|l|p{0.7\columnwidth}|}
\hline
\textbf{Notation} & \textbf{Meaning}                                                                                    \\ \hline
$L$                 & Desired summary length (number of words)                                                      \\ \hline
$n$                 & Number of sentences in the document   \\ \hline
$g$                 & Number of segments in the document 
\\ \hline
$m$                 & Number of distinct content words in the document  
\\ \hline
$i$                 & Index for sentence  ($i= [1 \dots n]$) \\ \hline
$j$                 & Index for content word    ($j = [1 \dots m]$)     
\\ \hline
$k$                 & Index for segment     ($k= [1 \dots g]$)   
\\ \hline
$x_i$              & Indicator variable for sentence $i$ (1 if sentence $i$ is to be included in summary, 0 otherwise) 
\\ \hline
$y_j$              & Indicator variable for content word $j$ (1 if $j$ is to be included in summary, 0 otherwise)        
\\ \hline
$p_i$ & Indicator variable for sentence $i$ citing a prior-case (1 if there is a citation, 0 otherwise) 
\\ \hline
$a_i$ & Indicator variable for sentence $i$ citing a statute (1 if there is a citation, 0 otherwise) 
\\ \hline 
$L(i)$              & Number of words in sentence $i$       
\\ \hline
$I(i)$              & Informativeness of sentence $i$    
\\ \hline
$C(i)$              & Set of content words in sentence $i$  
\\ \hline
$position(i)$     & Position of sentence $i$ in the document \\ \hline
$Score(j)$       & Score (a measure of importance) of content word $j$   
\\ \hline 
$T_j$              & Set of sentences where content word $j$ is present 
\\ \hline
$Weight (k)$ & Weight (a measure of importance) of segment $k$     
\\ \hline
$S_k$              & Set of sentences belonging to segment $k$   
\\ \hline
$NOS_k$         & Minimum number of sentences to be selected from segment $k$ in the summary   
\\ \hline
\end{tabular}

\caption{Notations used in the \algo algorithm}
\label{tab:symboltable}
\end{table}

We consider that a case document has a set of $g$ rhetorical segments (e.g., the $g=8$ rhetorical segments stated in Section~\ref{sub:rhet-roles}), and the summary is supposed to contain a representation of each segment (as indicated in~\cite{how-to-brief-case-cuny,intro-case-briefing-northwestern,case-notes-open-univ-UK}). 
The algorithm takes as input -- (i)~a case document where each sentence has a label signifying its rhetorical segment, and (ii)~the desired number of words $L$ in the summary. 
The algorithm then outputs a summary of at most $L$ words containing a representation of each segment.

\vspace{2mm}
\noindent \textbf{The optimization framework:}
We formulate the summarization problem using an optimization framework.
The ILP formulation maximizes the following factors:\\
\textbf{(i) Informativeness of a sentence}: The informativeness $I(i)$ of a sentence $i$ defines the importance of a sentence in terms of its information content. More informative sentences are likely to be included in the summary. We assume that domain experts will define how to estimate the informativeness of a sentence.\\
\textbf{(ii) Content words}: Content words signify domain-specific vocabulary (e.g., terms from a legal dictionary, or names of statutes, etc) and noun-phrases. 
The importance of each content word $j$ is given by  $Score(j)$.
We assume that domain experts will define what words/terms should be considered as `content words', and how important the content words are.  

A summary of length $L$ words, consisting of the most informative sentences and content words, is achieved by maximizing the following objective function:
\begin{equation}
\label{eqn:ilp}
    max ( \sum_{i=1}^{n} I(i) \cdot  x_i  + \sum_{j=1}^{m} Score(j) \cdot y_j  )
\end{equation}
subject to constraints 


\begin{equation}
\label{eqn:cons1}
    \sum_{i=1}^{n} x_i \cdot L(i) \leq L
\end{equation}


\begin{equation}
\label{eqn:cons2}
    \sum_{j \in C(i)} y_j >= \left | C(i) \right | \cdot x_i , \;\; j = [1 \dots m]
\end{equation}

\begin{equation}
\label{eqn:cons3}
    \sum_{i \in T_j} x_i \geq y_j , i=\left [ 1 \dots n \right ]
\end{equation}
\begin{equation}
\label{eqn:cons4}
    \sum_{i \in S_k} x_i \geq NOS_k , \;\; k= [1 \dots g]
\end{equation}



The objective function in Eqn.~\ref{eqn:ilp} tries to maximize the inclusion of informative sentences (through the $x_i$ indicator variables) and the number of important content words (through the $y_j$ indicator variables). 
Here $x_i$ (respectively, $y_j$) is set to $1$ if the algorithm decides that sentence $i$ (respectively, content word $j$) should be included in the summary.
Eqn.~\ref{eqn:cons1} constraints that the summary length is at most $L$ words (note from Table~\ref{tab:symboltable} that $L(i)$ is the number of words in sentence $i$). 
Eqn.~\ref{eqn:cons2} implies that if a particular sentence $i$ is selected for inclusion in the summary (i.e., if $x_i = 1$), then the all the content words contained in that sentence ($C(i)$) are also selected. 
Eqn.~\ref{eqn:cons3} suggests that if a content word $j$ is selected for inclusion in the summary (i.e., if $y_j = 1$), then at least one sentence where that content word in present is also selected. 

Eqn.~\ref{eqn:cons4} ensures from each segment $k$ a minimum number of sentences $NOS_k$ should be selected in the summary. This is a key step that ensures representation of all rhetorical segments in the generated summary. We assume that suitable values of $NOS_k$ will be obtained from domain knowledge.

Any ILP solver can be used to solve the ILP; we specifically used the GUROBI optimizer (\url{http://www.gurobi.com/}).
Finally, those sentences for which $x_i$ is set to $1$ will be included in the summary generated by \algo.

\vspace{2mm}
\noindent \textbf{Handling Redundancy:} \algo implicitly handles redundancy in the summary through the $y_j$ variables that indicate inclusion of content words (refer to Eqn.~\ref{eqn:cons2}). 
According to Eqn.~\ref{eqn:cons2}, if a sentence is included in the summary, then the content words in that sentence are also selected. 
Consider that two sentences $i$ and $i'$ are similar, so that including both in the summary would be redundant. 
The two sentences are expected to have the same content words. Hence, adding both $i$ and $i'$ (which have the same content words) will {\it not} help in maximizing the objective function. 
Hence the ILP method is expected to refrain from adding both sentences, thus prohibiting redundancy in the summary.

\vspace{2mm}
\noindent \textbf{Applying \algo to case documents of a specific jurisdiction:}
It can be noted that this section has given only a general description of \algo, where it has been assumed that many details will be derived from domain knowledge (e.g., the  informativeness or importance of a sentence or content word, or values of $NOS_k$). 
In the next section, we specify how these details are derived from the guidelines given by domain experts from India.

%% file: sections/guidelines-india.tex
In this section, we describe how \algo was adopted to summarize Indian case documents.
We first describe the summarization guidelines stated by law experts from India, and then discuss how to adopt those guidelines into \algo.
We then discuss comparative results of the proposed algorithm and the baselines on the India dataset.

\subsection{Guidelines from Law experts}
\label{sec:guidelines-india}

We consulted law experts -- senior law students and  faculty members from the Rajiv Gandhi School of Intellectual Property Law (a reputed law school in India) -- to know how Indian case documents should be summarized.
Based on the rhetorical segments described in Section~\ref{sub:rhet-roles}, the law experts suggested the following guidelines. \\
$\bullet$ {\bf (G1)} In general, the summary should contain representations from all segments of a case document, except the segment `Ruling by Lower Court' which may be omitted. 
\textbf{The relative importance of segments} in a summary should be: Final judgment $>$ Issue $>$ Fact $>$ (Statute, Precedent, Ratio) $>$ Argument.\\
$\bullet$ \textbf{(G2)} {\bf The segments `Final judgement' and `Issue' are exceptionally important for the summary}. The segments are usually very short in the documents, and so they can be included completely in the summary.\\
$\bullet$ \textbf{(G3)} The important sentences in various rhetorical segments (Fact, Statute, Precedent, Ratio) should be decided as follows -- 
\textbf{(a)} Fact: sentences that appear at the beginning; 
\textbf{(b)} Statute : sentences that contain citations to an Act; 
\textbf{(c)} Precedent : sentences that contain citation to a prior-cases; 
\textbf{(d)} Ratio : sentences appearing at the end of the document and sentences that contain citation to an act/law/prior-case.\\
$\bullet$ \textbf{(G4)} The summary must contain sentences that give important details of the case, including the Acts and sections of Acts that were referred, names of people, places etc. that concern the case, and so on. Also sentences containing specific legal keywords are usually important.

\subsection{Operationalizing the guidelines}

\begin{table*}[tb]
\renewcommand{\arraystretch}{1.2}
\begin{tabular}{|l|p{0.7\textwidth}|}

\hline
\textbf{Guideline}  & \textbf{Operationalization in \algo}                                                                                                                            \\ \hline
\textbf{(G1) Segment weights} & Highest $Weight(k)$ assigned to Final judgement and Issue; weight decreasing exponentially / linearly for other segments: $Final Judgement > Issue > Fact  > Statute, Ratio, Precedent  > Argument$
\\ \hline
\begin{tabular}[c]{@{}l@{}} \textbf{(G2) Segment} \\ \textbf{representation} \end{tabular}   & \begin{tabular}[c]{@{}l@{}}
$\begin{aligned}
NOS_k=
\begin{cases} 
\left | S_k \right |, \text{if}\ k=\text{Final\  judgement, Issue} \\
min (2, \left | S_k \right |),  \text{otherwise} 
\end{cases}
\end{aligned}$
  \end{tabular} \\ \hline                                
\begin{tabular}[c]{@{}l@{}} \textbf{(G3) Sentence} \\ \textbf{Informativeness}      \end{tabular}           &   
\begin{tabular}[c]{@{}l@{}}
$\begin{aligned}
I(i)=
    \begin{cases}
      Weight(k) \cdot (1/position(i)), \text{if}\ k=\text{Fact} \\
Weight(k) \cdot a_i, \text{if}\ k=\text{Statute} \\
Weight(k) \cdot p_i, \text{if}\ k=\text{Precedent} \\
      Weight(k) \cdot position(i) \cdot (p_i \text{ OR }  a_i), \text{if}\ k=\text{Ratio} \\
Weight (k), \text{otherwise} 
    \end{cases}
    \end{aligned}$
  \end{tabular}
\\ \hline
\begin{tabular}[c]{@{}l@{}} \textbf{(G4) Content Words} \& \\
\textbf{their weights} \end{tabular}  & \begin{tabular}[c]{@{}l@{}}
$\begin{aligned}
Score(j) = 
\begin{cases}
5, \text{if j = Act/sections of Acts (e.g.,Indian Penal Code, 1860; Article 15, etc.)} \\
3, \text{if j = Keywords from a legal dictionary}\\ 
1, \text{if j = Noun Phrases  = 1}
\end{cases}
    \end{aligned}$
\end{tabular}                                              \\ \hline
\end{tabular}
\caption{Parameter settings in \algo for operationalizing the guidelines provided by law experts for Indian case documents (stated in Section~\ref{sec:guidelines-india}).  Parameter settings such as weights of various segments, how to compute importance of sentences in various segments, etc. are decided based on the segment $k$. The symbols are explained in Table~\ref{tab:symboltable}.}
\label{tab:tuning}
\vspace{-5mm}
\end{table*}

Table~\ref{tab:tuning} shows how the above-mentioned guidelines from law experts in India have been operationalized in \algo.  

\vspace{1mm}
\noindent $\bullet$ {\bf Operationalizing G1:}
We experiment with two ways of assigning weights to segments -- linearly decreasing and exponentially decreasing -- based on guideline~G1 stated above. 
We finally decided to go with the exponentially decreasing weights.
Specifically, we assign weight $2^7$ to the `Final Judgement' segment (that is judged to be most important by experts), followed by $2^6$ for `Issue', $2^5$ for `Fact', $2^3$ for `Statute', `Ratio \& `Precedent' and finally $2^1$ for `Argument'.

\vspace{1mm}
\noindent $\bullet$ {\bf Operationalizing G2:}
As per guideline~G2, $NOS_k$ (the minimum number of sentences from a segment $k$) is set to ensure a minimum representation of every segment in the summary -- the `final judgement' and `issue' rhetorical segments are to be included fully, and at least $2$ sentences from every other segment are to be included.

\vspace{1mm}
\noindent $\bullet$ {\bf Operationalizing G3:}
The informativeness of a sentence $i$ defines the importance of a sentence in terms of its information content. More informative sentences are likely to be included in the summary. To find the informativeness $I(i)$ of a sentence $i$, we use the guideline~G3 stated above. $I(i)$ therefore depends on the rhetorical segment $k$ (Fact, Issue, etc.) that contains a particular sentence $i$. 
For instance, G3(a) dictates that sentences appearing at the beginning of the rhetorical segment `Fact' are important. We incorporate this guideline by weighing the sentences within `Fact' by the inverse of its position in the document. 

According to~G3(b) and~G3(c), sentences that contains mentions of Statutes/Act names (e.g., Section 302 of the Indian Penal Code, Article 15 of the Constitution, Dowry Prohibition Act 1961, etc.) and prior-cases/precedents are important. We incorporate this guideline through a Boolean/indicator variable $a_i$ which is $1$ if the sentence contains a mention of a Statute. We use regular expression patterns to detect Statute/Act name mentions. Similarly for detecting if a sentence contains a reference to a prior case, we use the Boolean variable $p_i$. If a regular expression pattern of the form \textit{Party 1 vs. Party 2} is found in a sentence $i$, $p_i$ is set to $1$. 

For understanding important Ratio sentences, we lookup guideline~G3(d), which mandates that sentences containing references to either a Statute/Act name or a prior-case are important to be included in the summary. We therefore use the two Boolean/indicator variables $a_i$ and $p_i$ for detecting the presence of a Statute/Act name or a prior-case. If any one of them occurs in the sentence, we consider that sentence from segment Ratio to be informative.

\vspace{1mm}
\noindent $\bullet$ {\bf Operationalizing G4:}
Based on guideline~G4 stated above, we need to identify some important `content words' whose presence would indicate the sentences that contain especially important details of the case. 
To this end, we consider three types of content words -- mentions of Acts/sections of Acts, keywords from a legal dictionary, and noun phrases (since names of people, places, etc. are important). 
For identifying legal keywords, we use a legal dictionary from the website \url{advocatekhoj.com}. 
For identifying mentions of Acts and statutes, we use a comprehensive list of Acts in the Indian judiciary, obtained from  Westlaw India (\url{westlawindia.com}).
As content word scores ($Score(j)$), we assign a weight of $5$ to statute mentions, $3$ to legal phrases and $1$ to noun phrases; these scores are based on suggestions by the legal experts on the relative importance of sentences containing different types of content words.

%% file: sections/baselines.tex
This section describes the baseline summarization methods that we consider for comparison with the proposed method. 
Also the experimental setup used to compare all methods is discussed.
\vspace{-1mm}
\subsection{Baseline summarization methods} \label{sub:baselines}

As described in Section~\ref{sec:related}, there are four classes of methods that can be applied for extractive summarization of legal case documents.  We consider some representative methods from each class as baselines.

\vspace{2mm}
\noindent \textbf{Unsupervised domain-independent methods:} We consider as baselines the following popular algorithms from this class: (1)~Luhn~\cite{Luhn:1958}, (2)~LexRank~\cite{Erkan:2004}, (3)~Reduction~\cite{jing2000sentence}, and (4)~LSA~\cite{Gong:2001} (see Section~\ref{sub:related-udi} for brief descriptions of these methods).

\vspace{2mm}
\noindent \textbf{Unsupervised domain-specific methods:}  Section~\ref{sub:related-uds} gave brief descriptions of these methods. Among such methods, we consider the following four as baselines -- 
(1)~LetSum~\cite{farzindar2004letsum},
(2)~KMM~\cite{saravanan2006improving},
(3) ~CaseSummarizer~\cite{polsley2016casesummarizer} and (4)~a simplified version of MMR~\cite{zhong2019automatic} --
in the absence of identical datasets, we adopt only the Maximum Margin Relevance module and use it to summarize a document.

\vspace{2mm}
\noindent \textbf{Supervised domain-specific methods:} 
From this class of algorithms, we consider Gist~\cite{liu2019extracting}, a recent method that applies general ML algorithms to the task. The best performance was observed by using Gradient Boosted Decision Tree as the ML classifier, which we report (see Section~\ref{sub:related-sds} for details of the method).

\vspace{2mm}
\noindent \textbf{Supervised domain-independent methods:} Among these methods, we consider the neural method SummaRuNNer~\cite{nallapati2017summarunner} (implementation available at \url{https://github.com/hpzhao/SummaRuNNer}). Similar to Gist, they consider the task of extractive summarization as binary classification problem.  The classifier returns a ranked list of sentences based on their prediction/confidence probabilities about their inclusion in the summary. We include sentences from this list in decreasing order of their predicted probabilities, until the desired summary length is reached. 

We also apply a recent BERT-based summarization method BERTSUM~\cite{bertsum} (implementation available at \url{https://github.com/nlpyang/PreSumm}). 
In BERTSUM, the sentence selection function is a simple binary classification task (whether or not to include a sentence in the summary).\footnote{The original BERTSUM model uses a post-processing step called \textit{Trigram Blocking} that excludes a candidate sentence if it has a significant amount of trigram overlap with the already generated summary (to minimize redundancy in the summary). However, we observed that this step leads to summaries that are too short, as also observed in~\cite{ExtendedSumm}. Hence we ignore this step.} Similar to SummaRuNNer, we use the ranked list of sentences based on the confidence probabilities of their inclusion in the summary. We include sentences one-by-one into the final summary, until the desired length is reached.

\vspace{2mm}
\noindent {\bf Implementations of the baselines:} 
For Luhn, LexRank, Reduction and LSA, we use the implementations from the Python \textit{sumy} module. 
The implementations of SummaRuNNEr and BERTSUM are also available publicly, at the URLs stated above.
We implemented the legal domain-specific summarization methods -- CaseSummarizer, LetSum, GraphicalModel, MMR and Gist -- following the descriptions in the corresponding papers as closely as possible (the same parameter values were used as specified in the papers).


\vspace{2mm}
\noindent \textbf{Converting abstractive summaries to extractive for training supervised summarization models}: 
The supervised methods Gist, SumaRuNNer and BERTSUM require labelled data for training, where every sentence in the source document must be labeled as $1$ if this sentence is suitable for inclusion in the summary, and labeled as $0$ otherwise.
The publicly available headnotes in our training datasets (see Section~\ref{sec:dataset-india}) are abstractive in nature and cannot be directly used to train such methods.  Hence we convert the abstractive summaries to extractive for training these  methods. 

To this end, we adopt the technique mentioned in~\cite{nallapati2017summarunner}. Briefly, we assign label $1$ (suitable for inclusion in the summary) to those sentences from the source document (full text), that greedily maximizes the ROUGE 2-overlap with the abstractive summary. The rest of the sentences in the source document are  assigned label $0$.

\subsection{Experimental setup} \label{sub:expt-setup}

Now we discuss the setup for our experiments to compare the performances of various summarization models.

\vspace{2mm}
\noindent \textbf{Target length of summaries}: Summarization models need to be told the target length $L$ (in {\it number of words}) of the summaries that they need to generated. 
For every document in the India dataset, we have two gold standard summaries written by two domain experts. For a particular document, we consider the target summary length $L$ to be the average of the number of words in the summaries (for that document) written by the two experts. 

For each document, every summarization model is made to generate a summary of length at most $L$ words, which ensures a fair comparison among the different models. 

\vspace{2mm}
\noindent \textbf{Evaluation of summary quality:} We use the standard ROUGE metrics for evaluation of summary quality (computed using \url{https://pypi.org/project/rouge/} v. 1.0.0). 
For a particular document, we have gold standard summaries written by two legal experts (as was described in Section~\ref{sec:dataset-india}). We compute the ROUGE scores of the algorithmic summary and each of the gold standard summary, and take the average of the two.
We remove a common set of English stopwords from both the gold standard summaries and the algorithm-generated summaries before measuring ROUGE scores.
We report ROUGE-2 Recall and F-score, and ROUGE-L Recall and F-scores. The scores are averaged over all $50$ documents in the evaluation set.

%% file: sections/results-india.tex
This section compares the performances of the proposed \algo and all the baseline algorithms. We compare the performances broadly in two ways -- 
(i)~how good the summaries for the full documents are, and 
(ii)~how well each individual rhetorical segment is summarized.
Table~\ref{tab:rouge-all-india} and Table~\ref{tab:rouge-segment-india} show the performances of various methods in summarizing the Indian Supreme Court case documents. In both tables, ROUGE scores are averaged over the $50$ documents in the India evaluation set.

\subsection{Evaluation of the full summaries}

\begin{table}[tb]
\renewcommand{\arraystretch}{1.1}

\begin{tabular}{|c|c|c|c|c|}
\hline
\multirow{2}{*}{\textbf{Algorithm}} & \multicolumn{2}{c|}{\textbf{ROUGE 2}}                                             & \multicolumn{2}{c|}{\textbf{ROUGE-L}}                            \\ \cline{2-5} 
                                    & \textbf{R}                              & \textbf{F}                              & \textbf{R}             & \textbf{F}                              \\ \hline
\multicolumn{5}{|c|}{\textbf{Unsupervised Domain Independent}}                                                                                                                             \\ \hline
(i) LexRank                             & 0.3662                                  & 0.3719                                  & 0.5068                 & 0.5392                                  \\ \hline
(ii) LSA                                 & 0.3644                                  & 0.3646                                  & 0.5632                 & 0.5483                                  \\ \hline
(iii) Luhn                                & 0.3907                                  & 0.3873                                  & 0.5424                 & 0.5521                                  \\ \hline
(iv) Reduction                           & 0.3778                                  & 0.3814                                  & 0.5194                 & 0.542                                   \\ \hline
\multicolumn{5}{|c|}{\textbf{Unsupervised Domain Specific}}                                                                                                                                \\ \hline
(v) LetSum                              & 0.403                                   & 0.4137                                  & 0.5898 & {\ul \textit{0.5846*}} \\ \hline
(vi) KMM                                 & 0.354                                   & 0.3546                                  & 0.5407                 & 0.5385                                  \\ \hline
(vii) CaseSummarizer                      & 0.3699                                  & 0.3765                                  & 0.4925                 & 0.5349                                  \\ \hline
(viii) MMR                                 & 0.3733                                  & 0.3729                                  & {\ul \textit{0.6064}}                & 0.568                                   \\ \hline
\algo (proposed) & \textbf{0.4323} & \textbf{0.4217}        & \textbf{0.6831}  & \textbf{0.6017}   \\
\hline
\multicolumn{5}{|c|}{\textbf{Supervised Domain Independent}}                                                                                                                               \\ \hline
(ix) SummaRuNNer                         & {\ul \textit{0.4104*}} & {\ul \textit{0.4149*}} & 0.5835                 & 0.5821                                  \\ \hline
(x) BERTSUM                             & 0.4044                                  & 0.4048                                  & 0.56                   & 0.5529                                  \\ \hline
\multicolumn{5}{|c|}{\textbf{Supervised Domain Specific}}                                                                                                                                  \\ \hline
(xi) Gist                                & 0.3567                                  & 0.3593                                  & 0.5712                 & 0.5501                                  \\ \hline
\end{tabular}
\caption{Performance of the summarization methods as per ROUGE metrics (R = recall, F = Fscore). All values are averaged over the $50$ documents in the evaluation set. The best value for each measure is in \textbf{bold} and the second best in \underline{underlined}. Entries with asterisk(*) indicates DELSumm is \textit{not} statistically significant than baselines using 95\% confidence interval by Student’s T-Test.} 
\label{tab:rouge-all-india}
\end{table}

Table~\ref{tab:rouge-all-india} shows the performances of summarizing the full documents in terms of ROUGE-2 and ROUGE-L Recall and F-scores. Considering all the approaches, \algo achieves the best performance across all the measures. The second-best metric values are achieved by LetSum (For ROUGE-L F-score), MMR (for ROUGE-L Recall), and SummaRuNNer (for ROUGE-2 Recall and F-score). To check if their performances are statistically significantly different from those achieved by \algo, we perform Student's T-Test at 95\% ($p<0.05$) with these three methods. Entries that are \textit{not} statistically significant are marked with asterisk (*) in Table~\ref{tab:rouge-all-india}.

It can be noted that, out of all the domain-specific methods in Table~\ref{tab:rouge-all-india} (both supervised and unsupervised), \algo performs the best for all the measures. 
While LetSum is the only prior work that considers rhetorical segments while generating the summary (see Section~\ref{sec:related}), none of these methods take into account domain-specific guidelines on what to include in the summary. 
\algo on the other hand models the legal knowledge about what to include in a summary, thus being able to perform better.

Importantly, \algo outperforms even supervised neural models such as SummaRuNNer and BERTSUM (though the difference in the performance of SummaRuNNer with respect to \algo is \textit{not} statistically significant in terms of ROUGE-2 scores).
Note that \algo is a completely unsupervised method while SummaRuNNer and BERTSUM are deep-learning based supervised methods trained over $7,100$ document-summary pairs. 
The fact that \algo still outperforms these supervised models is because \algo intelligently utilizes domain knowledge while generating the summaries.




\subsection{Evaluation of segment-wise summarization}

\begin{table}[tb]
\renewcommand{\arraystretch}{1.3}
\small
\scalebox{0.78}{
\begin{tabular}{|c|c|c|c|c|c|c|}
\hline
\textbf{Algorithm} & \textbf{\begin{tabular}[c]{@{}c@{}}Final\\ judgement\\ (2.9\%)\end{tabular}} & \textbf{\begin{tabular}[c]{@{}c@{}}Issue\\ (1.2\%)\end{tabular}} & \textbf{\begin{tabular}[c]{@{}c@{}}Facts\\ (23.9\%)\end{tabular}} & \textbf{\begin{tabular}[c]{@{}c@{}}Statute\\ (7.1\%)\end{tabular}} & \textbf{\begin{tabular}[c]{@{}c@{}}Precedent\\ +Ratio\\ (53.1\%)\end{tabular}} & \textbf{\begin{tabular}[c]{@{}c@{}}Argument\\ (8.6\%)\end{tabular}} \\ \hline
\multicolumn{7}{|c|}{\textbf{Unsupervised, Domain Independent}}                                                                                  \\ \hline
LexRank            & {\color[HTML]{FE0000} {\ul 0.0619}}                                          & 0.3469                                                           & 0.4550                                                            & {\color[HTML]{FE0000} {\ul 0.2661}}                                & 0.3658                                                                         & 0.4284                                                              \\ \hline
LSA                & {\color[HTML]{FE0000} {\ul 0.0275}}                                          & {\color[HTML]{FE0000} {\ul 0.2529}}                              & 0.5217                                                            & {\color[HTML]{FE0000} {\ul 0.2268}}                                & 0.3527                                                                         & 0.3705                                                              \\ \hline
Luhn               & {\color[HTML]{FE0000} {\ul 0.0358}}                                          & {\color[HTML]{FE0000} {\ul 0.2754}}                              & 0.5408                                                            & {\color[HTML]{FE0000} {\ul 0.2662}}                                & {\color[HTML]{FE0000} {\ul 0.2927}}                                            & 0.3781                                                              \\ \hline
Reduction          & {\color[HTML]{FE0000} {\ul 0.0352}}                                          & 0.3153                                                           & 0.5064                                                            & {\color[HTML]{FE0000} {\ul 0.2579}}                                & 0.3059                                                                         & {\color[HTML]{009901} \textbf{0.4390}}                              \\ \hline
\multicolumn{7}{|c|}{\textbf{Unsupervised, Domain Specific}}\\ \hline

LetSum             & {\color[HTML]{FE0000} {\ul 0.0423}}                                          & 0.3926                                                           & 0.6246                                                            & 0.3469                                                             & 0.3853                                                                         & {\color[HTML]{FE0000} {\ul 0.2830}}                                 \\ \hline
KMM                & 0.3254                                                                       & {\color[HTML]{FE0000} {\ul 0.2979}}                              & 0.4124                                                            & 0.3415                                                             & 0.4450                                                                         & 0.416                                                               \\ \hline
CaseSummarizer     & {\color[HTML]{FE0000} {\ul 0.2474}}                                                                       & 0.3537                                                           & 0.4500                                                            & {\color[HTML]{FE0000} {\ul 0.2255}}                                & 0.4461                                                                         & 0.4184                                                              \\ \hline
MMR                & 0.4378                                                                       & 0.3548                                                           & 0.4442                                                            & {\color[HTML]{FE0000} {\ul 0.2763}}                                & 0.4647                                                                         & 0.3705                                                              \\ \hline
\algo    & {\color[HTML]{009901} \textbf{0.7929}}                                       & {\color[HTML]{009901} \textbf{0.6635}}                           & 0.5539                                                            & {\color[HTML]{009901} \textbf{0.4030}}                             & 0.4305                                                                         & 0.4370                                                              \\ \hline
\multicolumn{7}{|c|}{\textbf{Supervised, Domain Independent}}                                                                                                                                                                                                                                                                                                                                                                                                        \\ \hline
SummaRuNNer        & 0.4451                                                                       & {\color[HTML]{FE0000} {\ul 0.2990}}                              & 0.5231                                                            & {\color[HTML]{FE0000} {\ul 0.1636}}                                & {\color[HTML]{009901} \textbf{0.5215}}                                         & 0.3090                                                              \\ \hline
BERTSUM            & {\color[HTML]{FE0000} {\ul 0.0662}}                                          & 0.3544                                                           & {\color[HTML]{009901} \textbf{0.6376}}                            & {\color[HTML]{FE0000} {\ul 0.2535}}                                & 0.3121                                                                         & 0.3262                                                              \\ \hline
\multicolumn{7}{|c|}{\textbf{Supervised, Domain Specific}}                                                               \\ \hline
Gist               & 0.5844                                                                       & 0.3856                                                           & 0.4621                                                            & {\color[HTML]{FE0000} {\ul 0.2759}}                                & 0.4537                                                                         & {\color[HTML]{FE0000} {\ul 0.2132}}                                 \\ \hline
\end{tabular}
}
\caption{Segment-wise performance (ROUGE-L F-scores) of the methods. All values are averaged over the $50$ documents in the evaluation set. Values in the column headings are the \% of sentences in the full document that belong to a segment.  Values $< 0.3$ highlighted in red-underlined. The best value for each segment in green-bold.}
\label{tab:rouge-segment-india}
\vspace{-5mm}
\end{table}

\textit{Overall} ROUGE scores are not the best metrics for evaluating case document summaries. Law experts opine that, even methods that achieve high \textit{overall} ROUGE scores may not represent every segment well in the summary~\cite{bhattacharya2019comparative}. 
A segment-wise performance evaluation is practically important since law practitioners often intend to read the summary of a particular segment, and not always the full summary. 
Hence we perform a segment-wise performance evaluation of the algorithms. 

To this end, we proceed as follows. For each rhetorical segment (e.g., Facts, Issues), we extract the portions of an algorithmic  summary and the gold standard summary that represent the given segment, using the gold standard rhetorical labels of individual sentences (which are present in our evaluation set, as detailed in Section~\ref{sec:dataset-india}). 
Then we compute the ROUGE scores on those specific text portions only. 
For instance, to compute the ROUGE score on the `Fact' segment of a particular document, we only consider those sentences in the gold standard summary and the algorithmic summary, which have the rhetorical label `Fact' in the said document.
We report the average ROUGE score for a particular segment, averaged over all $50$ documents in the evaluation set.

Table~\ref{tab:rouge-segment-india}  shows the \textit{segment-wise} ROUGE-L F-scores (averaged over all $50$ documents in the evaluation set). The values $< 0.3$ are underlined and highlighted in red color, while the best value for each segment is highlighted in boldface and green color.

Similar to~\cite{bhattacharya2019comparative}, we also observe that many of the baseline methods could {\it not} represent the `Final judgement' and `Issue' well in their summaries. This flaw is especially critical since  these two segments are the most important in the summary according to the law experts (see Section~\ref{sec:guidelines-india}).
In contrast, \algo achieves a very high performance on these two important segments. 
This difference in performance is possibly because {\it these two segments are also the shortest segments} (constitutes only 2.9\% and 1.2\% of the whole document, as stated in the first row of Table~\ref{tab:rouge-segment-india}), and hence are missed by other methods which do {\it not} know of their domain-specific importance.
This observation show the necessity of an informed algorithm that can incorporate domain knowledge from experts. 

\algo also represents the `Statute' segment better than all other methods. For the `Statute' segment, the algorithm was formulated in such a way (through the $a_i$ variable) that it is able to incorporate sentences that contain mention of an Act/law. Other methods did not perform well in this aspect. 
The performance of \algo for the `Argument' segment is second-best ($0.4370$) after that of Reduction ($0.4390$). These values are very close, and the difference is \textit{not statistically significant}.

Our method is unable to perform as well for the `Precedent $+$ Ratio' and `Facts' segments as some other methods. 
Note that, these segments accounts for {\it maximum number of sentences in a document} and also forms a large part of the summary (see Table~\ref{tab:rouge-segment-india}, first row). Hence neural methods (e.g., BERTSUM) and methods relying of TF-IDF measures (e.g., LetSum) obtained relatively large amounts of training data for these segments, and hence performed well for these segments. 

Finally, although LetSum is a segment-aware algorithm, its \textit{summarization} mechanism is not strong enough to understand the importance of sentences in all segments; for instance, it performs very poorly for the `Final judgement' segment. DL methods fail to perform well for the smaller segments (e.g., Final judgement, Issue) for which lesser amounts of training data is available.

Overall, it is to note that \algo achieves a 
\textbf{much more balanced representation of segments in the summary}, compared to all the baseline methods. 
Contrary to the baseline methods, \algo shows decent performance for \textit{all} the segments. This has been taken care mainly through the constraint in Eqn.~\ref{eqn:cons4}. Hence, even when the most important segments (e.g., Final judgement) are optimized in the summary, the less important ones (e.g., Arguments) are {\it not} missed out.

\subsection{Discussion on the comparative results}


The results stated above suggest that our proposed method performs better than many existing domain-independent as well as legal domain-specific algorithms. 
The primary reason for this superior performance of \algo is that it is able to incorporate legal domain knowledge much more efficiently through a theoretically grounded approach. 
By doing this, it is able to surpass the performance of deep learning and machine learning approaches such as SummaRuNNer, BERTSUM and Gist (that are trained over $7,100$ document-summary pairs). 
These results show that if legal domain knowledge can be used intelligently, even simple unsupervised summarization algorithms can compete with state-of-the-art supervised learning approaches.

It can be argued that $7,100$ document-summary pairs are too less to train a deep learning summarization model; additionally, as stated in Section~\ref{sec:dataset-india}, the summaries (headnotes) in the training set may not be of the same quality as those in the evaluation set.
However, in domains such as legal, it is difficult and expensive to obtain rich quality training data in huge amounts. In such scenarios, \algo, being an unsupervised method, can work efficiently by incorporating domain knowledge.

Another advantage of \algo is that it is easily customizable -- if the requirement is that a particular segment should have more representation in the summary than the others, the segment weights and representations can be tuned accordingly. If the sentence importance measures are different, that too can be adjusted. 
Other domain-specific and supervised learning methods do not exhibit such flexibility.

%% file: sections/noisy-labels.tex
\algo requires sentences of the source document (full text) to be labelled with a rhetorical role prior to summarization. To this end, till now we have been using expert-annotated labels. 
However, in a real scenario, it is not possible to get expert-annotations for rhetorical role labelling for every document. 
A more practical approach is to first computationally generate these rhetorical labels through some algorithm that detects rhetorical roles of sentences in a legal document, and then consider the labelled document as input to \algo for summarization. 

Note that algorithmically derived rhetorical labels are likely to be noisy (i.e., not 100\% accurate). In this section, we analyse how the performance of \algo is affected by such noisy {\it algorithmic labels} when compared to gold standard labels. To this end, we perform the following experiment (which is inspired by the idea of $k$-fold cross validation that is popular in Machine Learning literature).

\vspace{2mm}
\noindent \textbf{Experimental Design:} Assume that we have a dataset $D$ consisting of $n$ documents, each having sentence-wise rhetorical label annotations as well as gold-standard summaries (e.g., the Indian dataset we are using in this paper, with $n=50$). 
We divide the dataset into $k$ folds (e.g., $k=5$). In each fold $f$, there are $n/k$ documents for evaluation, which we call as the evaluation set $E_f$. The remaining $n - (n/k)$ documents are considered for training a model for rhetorical role labeling, termed as training set $T_f$. 
We also have the expert annotated labels for the documents in fold $f$, and their gold standard summaries.  
For each fold, we use $T_f$ for training the state-of-the-art rhetorical role identification model Hier-BiLSTM-CRF from our prior work~\cite{bhattacharya2019identification}.\footnote{Implementation available at \url{https://github.com/Law-AI/semantic-segmentation}.} 

Using the trained model for that fold, we infer the rhetorical labels of the documents in the evaluation set $E_f$ for that fold. 
Now, we have two sets of rhetorical labels for the sentences in the documents in $E_f$ --
(i)~gold standard labels given by law experts, and (ii)~algorithmic labels assigned by the model~\cite{bhattacharya2019identification}.
We next summarize every document in $E_f$ with \algo twice, once using the gold standard sentence labels, and again using the algorithmic sentence labels. We perform three evaluations:

\noindent \textbf{(i) Accuracy of rhetorical labelling:} we check what fraction of the algorithmic labels match with the gold standard labels (over the documents in $E_f$). This measure gives an idea of the accuracy of the rhetorical labeling model~\cite{bhattacharya2019identification}.

\noindent \textbf{(ii) ROUGE scores with Gold Std. Labels:} We compute the ROUGE scores between the gold standard summaries and the summaries generated by \algo considering the gold standard labels. We average the ROUGE scores over all documents in the set $E_f$ (this is very similar to what we did in Section~\ref{sec:india-expts}).

\noindent \textbf{(iii) ROUGE scores with Algorithmic Labels:} We compute the ROUGE scores between the gold standard summaries and summaries generated by \algo with the algorithmic labels, i.e., labels inferred by the model trained over $T_f$. Again we average the ROUGE scores over  all documents in the set $E_f$.  

\noindent The differences between these two sets of ROUGE scores can be used to quantify the degradation in performance of \algo while using noisy algorithmic labels instead of gold standard labels.

\begin{figure}[tb]
\centering
  \includegraphics[width=0.8\columnwidth]{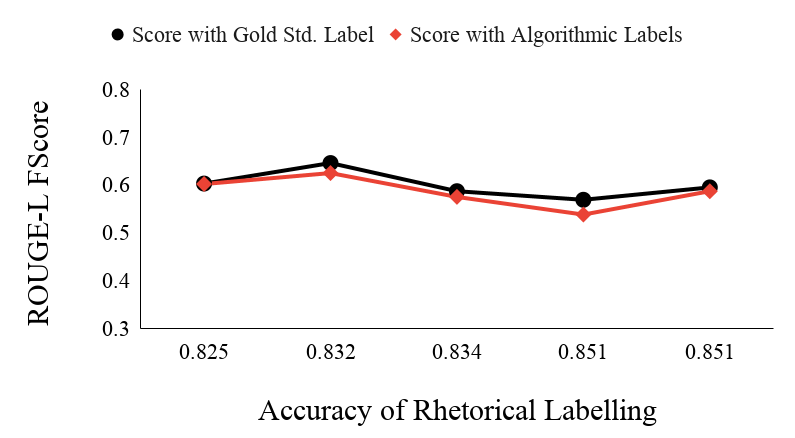}
\caption{Effect on the performance of \algo with algorithmic labels. There is minimal degradation of ROUGE scores when using algorithmic labels}
\label{fig:graph}
\end{figure}

\vspace{2mm}
\noindent \textbf{Results:} 
We perform the above experiment with the India dataset where $n = 50$, and we consider $k = 5$ folds.
Hence in each fold, we use a training set of $40$ documents and an evaluation set of $10$ documents. 
Figure~\ref{fig:graph} shows the results of the above experiment. The horizontal axis shows the rhetorical labeling accuracy across the $k=5$ folds; we see that the model~\cite{bhattacharya2019identification} achieves accuracies in the range $[0.825, 0.851]$, i.e., about 82\%--85\% of the algorithmic labels are correct. 
The figure shows the two types of ROUGE-L F-scores as described above -- one set using the algorithmic labels (denoted by red diamonds), and the other set using the gold-standard labels (denoted by black circles) -- for each of the $5$ folds. Each value is averaged over the $10$ documents in the evaluation set of the corresponding fold.
We find that the two sets of ROUGE scores are very close. In other words, even when noisy/inaccurate labels are used in \algo, the drop in performance is very low, indicating the robustness of \algo.

\begin{table}[tb]
\begin{tabular}{|c|c|c|c|c|}
\hline
\multirow{2}{*}{\textbf{Algorithm}} & \multicolumn{2}{c|}{\textbf{ROUGE-2}}                                    & \multicolumn{2}{c|}{\textbf{ROUGE-L}}                                                   \\ \cline{2-5} 
                                    & \textbf{R}                                 & \textbf{F}                  & \textbf{R}                                 & \textbf{F}                                 \\ \hline
LetSum                              & 0.4030                                     & 0.4137                      & 0.5898                                     & 0.5846                                     \\ \hline
MMR                                 & 0.3733                                     & 0.3729                      & 0.6064                                     & 0.5680                                     \\ \hline
SummaRuNNer                         & 0.4104                                     & {\ul \textit{0.4149}}       & 0.5835                                     & 0.5821                                     \\ \hline
DELSumm (GSL)                       & \textbf{0.4323}                            & \textbf{0.4217}             & \textbf{0.6831}                            & \textbf{0.6017}                            \\ \hline
\multicolumn{1}{|l|}{DELSumm (AL)}  & \multicolumn{1}{l|}{{\ul \textit{0.4193}}} & \multicolumn{1}{l|}{0.4075} & \multicolumn{1}{l|}{{\ul \textit{0.6645}}} & \multicolumn{1}{l|}{{\ul \textit{0.5897}}} \\ \hline
\end{tabular}
\caption{Performance of \algo with Gold Standard labels (GSL) and Algorithmic labels (AL) over the $50$ documents. Apart from the last row, all other rows are repeated from Table~\ref{tab:rouge-all-india}. While \algo (AL) shows slightly degraded performance than \algo (GSL), it still outperforms the other methods according to all ROUGE measures except ROUGE-2 F-score.}
\label{tab:rouge-all-india-crossval}
\vspace{-4mm}
\end{table}

Finally, we compute the ROUGE scores averaged over all $n=50$ documents across all $k=5$ folds. 
Table~\ref{tab:rouge-all-india-crossval} shows the average ROUGE scores achieved by \algo using the algorithmic labels (denoted `\algo AL'). 
For ease of comparison, we repeat some rows from Table~\ref{tab:rouge-all-india} -- the performance of \algo with gold standard rhetorical labels (denoted `\algo GSL'), and the performance of the closest competitors LetSum, MMR and SummaRuNNer. 
We observe that, even when \algo is used with automatically generated algorithmic labels (with about 15\% noise), it still performs better  than LetSum, MMR and SummaRuNNer (which were its closest competitors) according to all ROUGE measures except ROUGE-2 Fscore.
This robustness of \algo is because, through the sentence informativeness and content words measures in the ILP formulation, \algo can capture useful information even if the rhetorical labels may be a little inaccurate.

%% file: sections/conclusion.tex
We propose \algo, an unsupervised algorithm that systematically incorporates domain knowledge for extractive summarization of legal case documents. Extensive experiments and comparison with as many as eleven baselines, including deep learning-based approaches as well as domain-specific approaches, show the utility of our approach. 
The strengths our approach are: 
(i)~\algo systematically encodes domain knowledge necessary for legal document summarization into a computational approach, 
(ii)~although an unsupervised approach, it performs at par with supervised learning models trained over huge amounts of training data, 
(iii)~it is able to provide a summary that has a balanced representation from all the rhetorical segments, which is highly lacking in prior  approaches 
(iv)~inaccuracy in labels do not degrade the performance of \algo much, thus showing its robustness and rich information identification capabilities, and
(v)~the method is flexible and generalizable to summarize documents from other jurisdictions; all that is needed are the expert guidelines for what to include the summary, and how to identify important content words. 
The objective function and constraints can be adjusted as per the requirements of different jurisdictions (e.g., giving more weight to certain segments).
The implementation of \algo is publicly available at \url{https://github.com/Law-AI/DELSumm}.

In future, we plan to apply \algo to documents from other jurisdictions as well as to generate different types of summaries (e.g., for different stakeholders) and analyse the performance. 